\documentclass[letterpaper, 10 pt, conference]{ieeeconf}  

\IEEEoverridecommandlockouts                              

\usepackage{algorithm}
\usepackage{amsmath}
\usepackage{amsfonts}
\usepackage{amssymb}
\usepackage{algpseudocode}
\usepackage[caption=false,font=footnotesize]{subfig}
\usepackage{adjustbox}
\usepackage{comment}
\usepackage{xcolor}
\usepackage{newfloat}
\usepackage{listings}
\usepackage{soul}
\usepackage{pifont}
\newcommand{\cmark}{\ding{51}}
\newcommand{\xmark}{\ding{55}}
\usepackage{multirow}
\usepackage{booktabs}
\usepackage{hyperref}
\usepackage{booktabs,tabularx,array,makecell,graphicx}
\newcolumntype{Y}{>{\centering\arraybackslash}X}                
\newcolumntype{L}[1]{>{\raggedright\arraybackslash}p{#1}} 

\usepackage[caption=false,font=footnotesize]{subfig}

\title{ \bf
GRACE: A Unified 2D Multi-Robot Path Planning Simulator \& Benchmark for Grid, Roadmap, And Continuous Environments
}

\author{Chuanlong Zang$^{1,2}$, Anna Mannucci$^{1}$, Isabelle Barz$^{1}$, Philipp Schillinger$^{1}$, Florian Lier$^{1}$, Wolfgang Hönig$^{2}$
\thanks{$^{1}$Robert Bosch GmbH, Corporate Research, Stuttgart, Germany.
{\tt\small Chuanlong.Zang@de.bosch.com}}%
\thanks{$^{2}$Technical University of Berlin, Berlin, Germany.}%
\thanks{Code: \url{https://github.com/boschresearch/GRACE}}
}

\begin{document}

\maketitle
\thispagestyle{empty} 
\pagestyle{empty}
\bstctlcite{mybstctl} 


\begin{abstract}

Advancing Multi-Agent Pathfinding (MAPF) and Multi-Robot Motion Planning (MRMP) requires platforms that enable transparent, reproducible comparisons across modeling choices. Existing tools either scale under simplifying assumptions (grids, homogeneous agents) or offer higher fidelity with less comparable instrumentation. We present \emph{GRACE}, a unified 2D simulator+benchmark that instantiates the same task at multiple abstraction levels (grid, roadmap, continuous) via explicit, reproducible operators and a common evaluation protocol. Our empirical results on public maps and representative planners enable commensurate comparisons on a shared instance set. Furthermore, we quantify the expected representation–fidelity trade-offs (MRMP solves instances at higher fidelity but lower speed, while grid/roadmap planners scale farther). By consolidating representation, execution, and evaluation, GRACE thereby aims to make cross-representation studies more comparable and provides a means to advance multi-robot planning research and its translation to practice.

\end{abstract}

\section{Introduction} \label{introduction}

As autonomous systems become increasingly prevalent, coordinating multiple robots in shared environments is critical for unlocking their full potential. This challenge centers on finding collision-free paths or trajectories for agents in a known environment, while optimizing objectives such as makespan or sum-of-costs (SoC). This task is addressed by research areas including Multi-Agent Pathfinding (MAPF) and Multi-Robot Motion Planning (MRMP) \cite{stern_multi-agent_2019}. Typically, MAPF has focused on discrete-time grid-based formulations with homogeneous agents, allowing scalability and algorithmic guarantees \cite{okumura_engineering_2024, ren_map_2024}. MRMP operates in continuous space–time with kinematic/kinodynamic constraints and realistic footprints, capturing operational fidelity \cite{debord_trajectory_2018, gao_review_2024}.

Both views are valuable: discrete abstractions enable rapid design-space exploration and algorithmic benchmarking, while continuous models expose execution-level feasibility and performance limits. To advance research and ensure reliable deployment, fair and reproducible comparisons across these representations are crucial. 

The current ecosystem fragments this goal. Grid-centric MAPF simulators and benchmarks excel in large-scale comparisons but often omit agent dynamics, continuous time, and heterogeneity, limiting conclusions about real robots \cite{skrynnik_pogema_2025}. In contrast, general-purpose robotics simulators provide high fidelity, but make cross-representation and cross-planner benchmarking cumbersome: identical tasks are hard to express at multiple abstraction levels; and instrumentation is inconsistent between methods \cite{schaefer_benchmark_2023}. This fragmentation complicates answering basic questions such as: \emph{When is a roadmap abstraction good enough? When do roadmap or continuous models materially change conclusions? How do MAPF and MRMP compare on an equal footing under the same tasks and metrics?}

\begin{figure}[t]
\centering
\subfloat[Grid MAPF\label{fig:sub1}]{
  \includegraphics[width=0.3\linewidth]{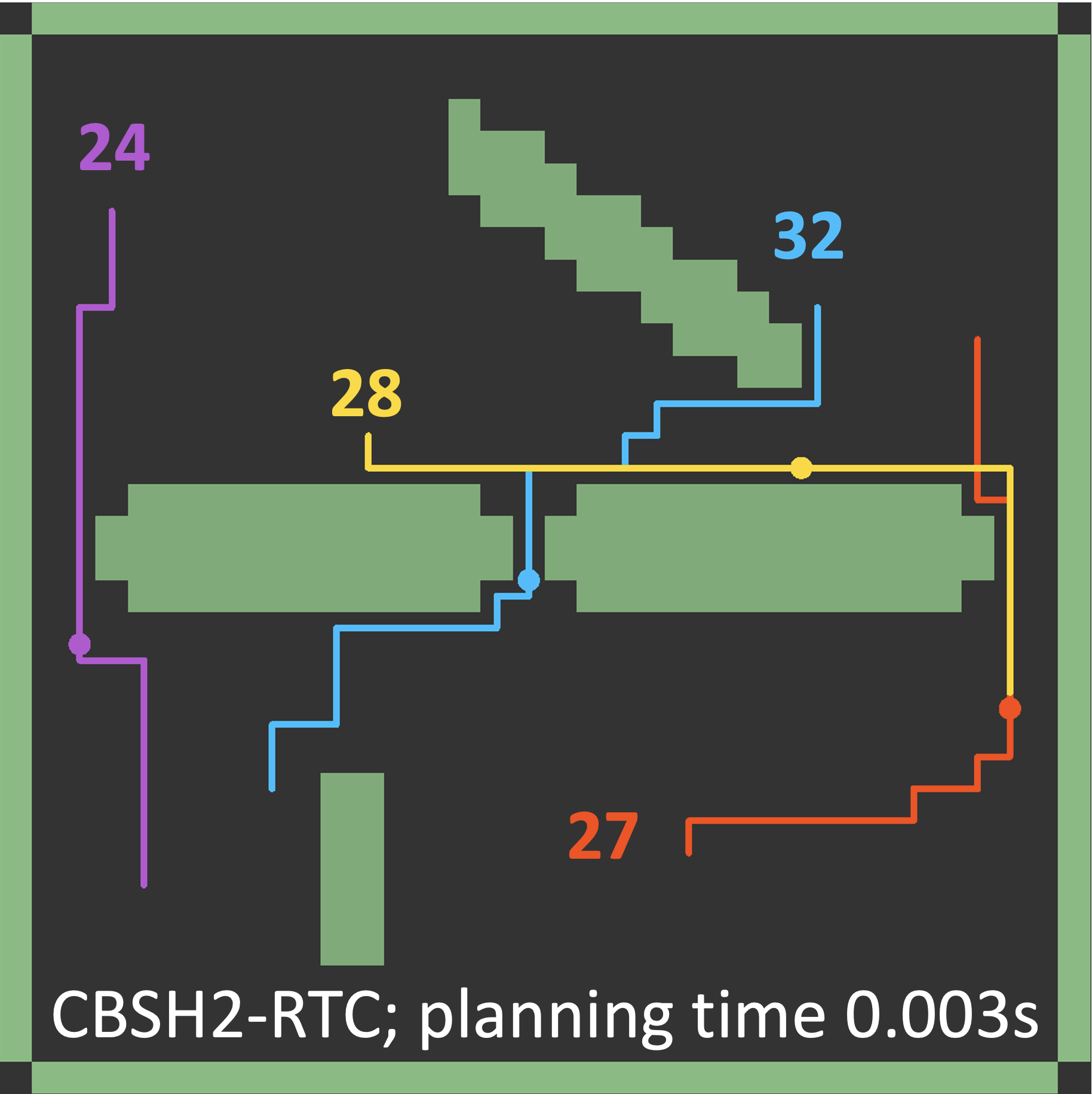}
}
\hfil
\subfloat[Roadmap MAPF\label{fig:sub2}]{
  \includegraphics[width=0.3\linewidth]{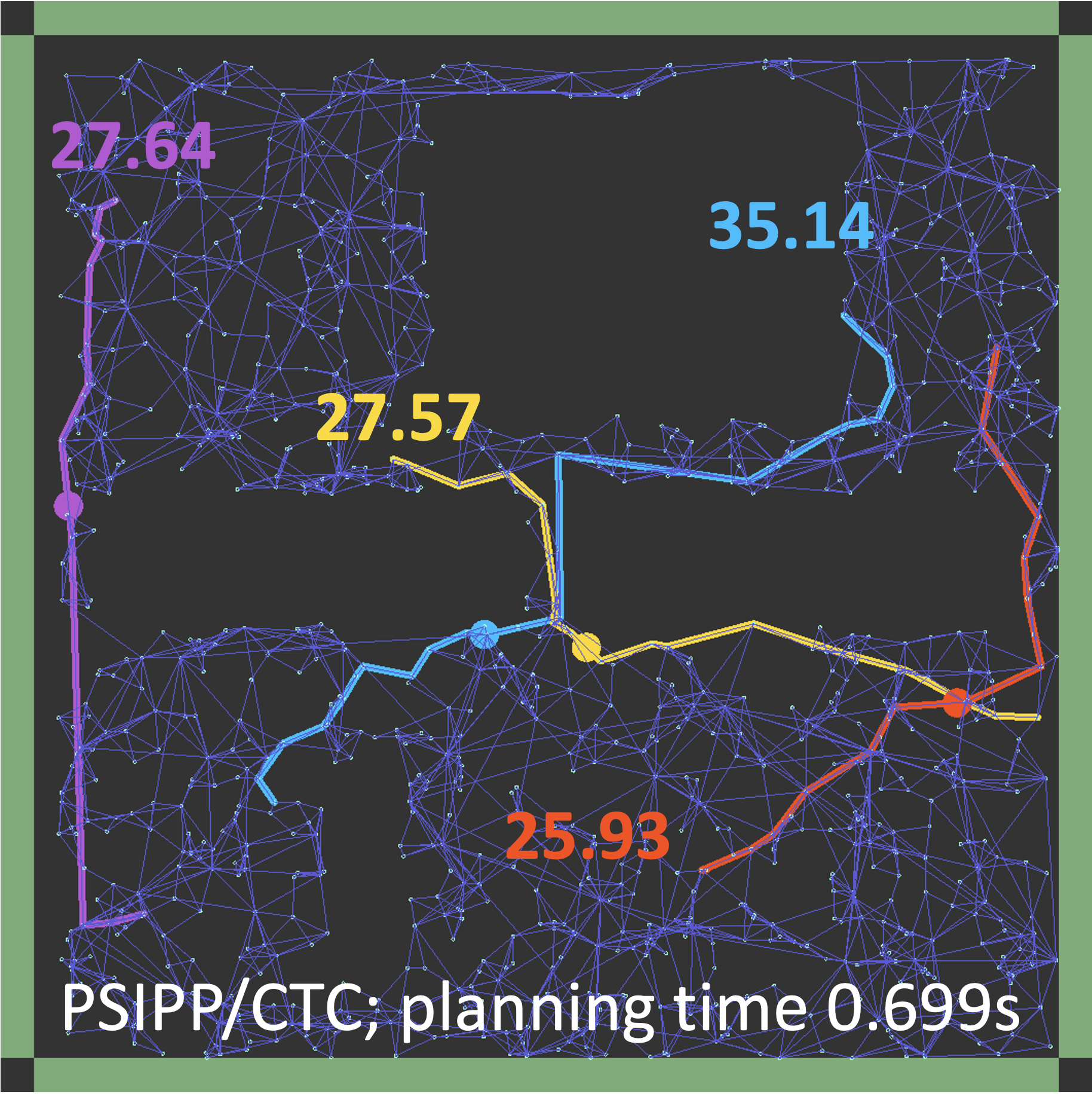}
}
\hfil
\subfloat[MRMP\label{fig:sub3}]{
    \includegraphics[width=0.3\linewidth]{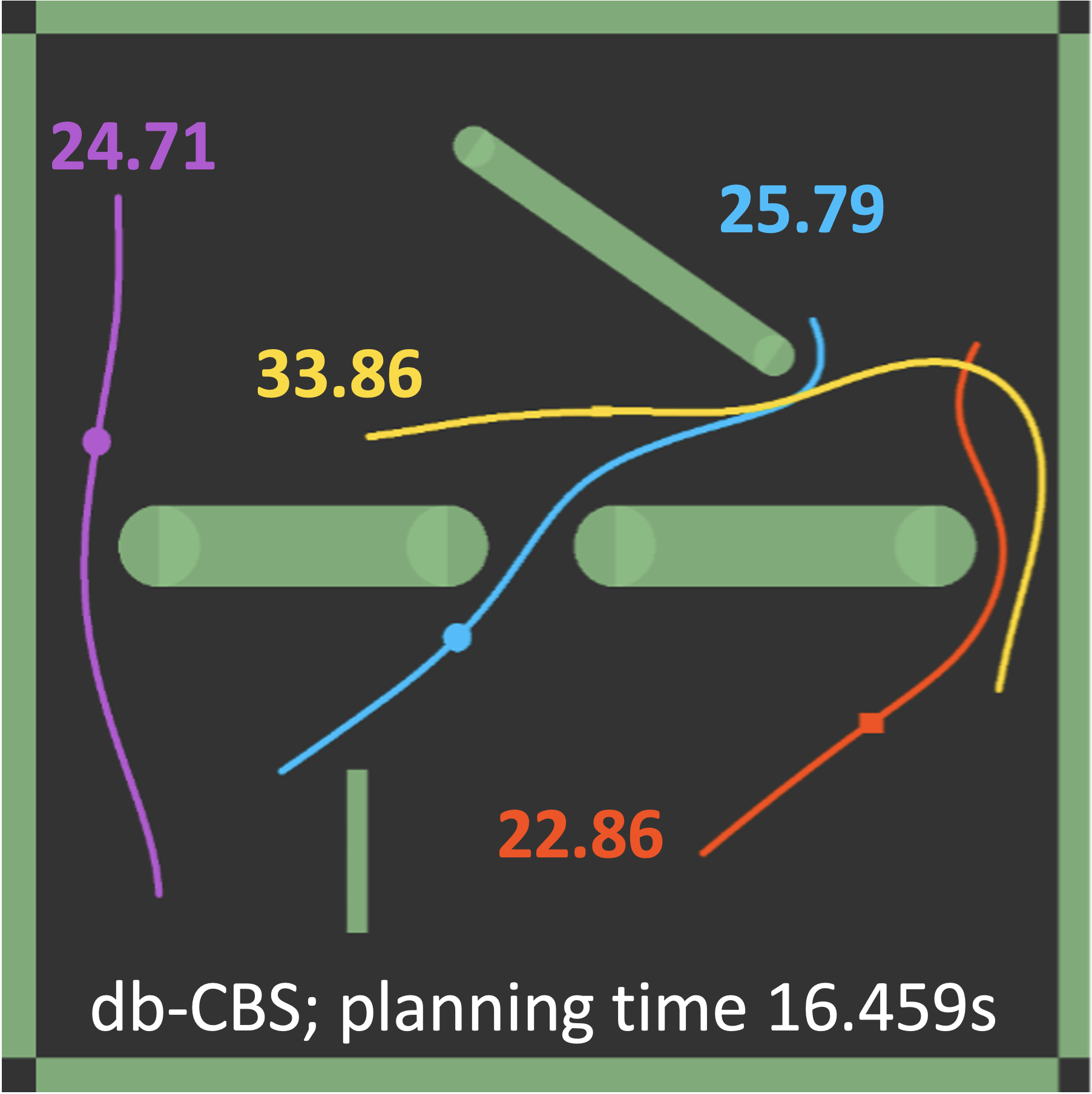}
}
\caption{\textbf{GRACE: Single scenario, multiple planning representations.} Identical agents/scenarios, distinct trajectories from representation-specific constraints: (a) occupancy-grid (4-connected discrete actions), (b) graph abstraction (single integrator), and (c) continuous space (double integrator, heterogeneous footprint). Individual path lengths highlight these representation-specific constraints.
}
\label{fig:top}
\end{figure}

This paper introduces GRACE, a unified 2D simulation and benchmarking platform designed to address this critical gap. GRACE uniquely supports grid, roadmap, and continuous environment models within a single API and evaluation protocol, enabling consistent evaluation across abstraction levels (as demonstrated in Fig.~\ref{fig:top}). Its flexible design allows seamless integration and comparative analysis of diverse MAPF and MRMP planners (classical, learning-based, or hybrid) under standardized scenario definitions and metrics.

\textbf{Contributions.} 1) GRACE, a unified simulator and benchmark for grid/roadmap/continuous planning, 
which systematically defines and implements explicit abstraction operators and cross-representation conversion mechanisms to ensure reproducible abstractions and standardized evaluation; and 2) an empirical study using GRACE that (i) compares representations on shared instance set, (ii) benchmarks planner families on grids, and (iii) reports determinism and observed scalability (up to \(\sim\)2k agents in our setup), highlighting representation–fidelity trade-offs.


\begin{table}[t]
\caption{Comparison of 2D multi-agent path planning platforms.}
\label{tab:benchmark}
\centering
\begingroup
\setlength{\tabcolsep}{3pt} 
\renewcommand{\arraystretch}{1.1}
\footnotesize
\begin{tabularx}{\columnwidth}{lYYYYYY}
\toprule
\textbf{Platform} & \textbf{Rep.} & \textbf{CST Env.} & \textbf{CST Plan.} & \textbf{Dyn.} & \textbf{Het.} & \textbf{Scale} \\
\midrule
\makecell[l]{MovingAI-related\\ \scriptsize \cite{sturtevant_benchmarks_2012,kaduri_experimental_2021,shen_tracking_2023,tan_benchmarking_2024}} &
G &
\textcolor{lightgray}{\xmark{}} &
\textcolor{lightgray}{\xmark{}} &
\textcolor{lightgray}{\xmark{}} &
\textcolor{lightgray}{\xmark{}} &
1k \\
\midrule
asprilo \cite{gebser_experimenting_2018} & 
G &
\textcolor{lightgray}{\xmark{}} & \textcolor{lightgray}{\xmark{}} & 
\textcolor{lightgray}{\xmark{}} & \textcolor{lightgray}{\xmark{}} &
50 \\
\midrule
Flatland \cite{mohanty_flatland-rl_2020} &
G &
\textcolor{lightgray}{\xmark{}} & \textcolor{lightgray}{\xmark{}} &
\textcolor{lightgray}{\xmark{}} & \textcolor{lightgray}{\xmark{}} &
6k \\
\midrule
LoRR \cite{chan_league_2024} &
G &
\textcolor{lightgray}{\xmark{}} & \textcolor{lightgray}{\xmark{}} &
\textcolor{lightgray}{\xmark{}} & \textcolor{lightgray}{\xmark{}} &
10k \\
\midrule
POGEMA \cite{skrynnik_pogema_2025} &
G &
\textcolor{lightgray}{\xmark{}} & \textcolor{lightgray}{\xmark{}} &
\textcolor{lightgray}{\xmark{}} & \textcolor{lightgray}{\xmark{}} &
1M \\
\midrule
MRP-Bench \cite{schaefer_benchmark_2023} &
C &
\cmark{} & \textcolor{lightgray}{\xmark{}} &
\textcolor{lightgray}{\xmark{}} & \textcolor{lightgray}{\xmark{}} &
10 \\
\midrule
mrmg-planning \cite{hartmann_benchmark_2025} &
C &
\cmark{} & \cmark{} & \cmark{} & \cmark{} &
10 \\
\midrule
REMROC \cite{heuer_benchmarking_2024} &
G+C &
\cmark{} & \cmark{} &
\cmark{} & \textcolor{lightgray}{\xmark{}} &
10 \\
\midrule
SMART \cite{yan_advancing_2025} &
G,C &
\cmark{} & \textcolor{lightgray}{\xmark{}} &
\cmark{} & \textcolor{lightgray}{\xmark{}} &
2k \\
\midrule
\textbf{GRACE (Ours)} &
G,R,C &
\cmark{} & \cmark{} & \cmark{} & \cmark{} &
2k \\
\bottomrule
\end{tabularx}

\par\smallskip
\textbf{Rep.} $\in\{$G=grid,R=roadmap,C=continuous$\}$: see Sec.~\ref{sec:env_model_ps}; G+C: hierarchical global + local planner. \textbf{CST Env./Plan.}=continuous space–time environment and planning; \textbf{Dyn.}=agent dynamics; \textbf{Het.}=heterogeneous agents; \textbf{Scale}=max number of agents in one experiment.

\endgroup
\end{table}

\section{Related Work}

This section reviews efforts to solve, simulate and benchmark MAPF and MRMP across grid, roadmap, and continuous representations. We first discuss existing benchmarks and platforms (Sec. \ref{existing_benchmarks}), followed by a survey of planner families across those representations (Sec. \ref{existing_algorithms}), as algorithmic assumptions critically influence fair benchmark evaluation.

\subsection{Benchmarks and Platforms in MAPF and MRMP}
\label{existing_benchmarks}

\begin{table}[t]
\caption{Comparison of 2D multi-agent path planning approaches.}
\label{tab:algorithm}
\centering
\begingroup
\setlength{\tabcolsep}{3pt} 
\renewcommand{\arraystretch}{1.1}
\footnotesize
\begin{tabularx}{\columnwidth}{lYYYYYYY}
\toprule
 \textbf{Planners} & \textbf{Rep.} & \textbf{Dyn.} & \textbf{Het.} & \textbf{Opt.}\textsuperscript{\dag} & \textbf{Comp.} & \textbf{OSS} & \textbf{Eval.} \\
\midrule
CBSH2-RTC \cite{walker_extended_2018} & G &
\textcolor{lightgray}{\xmark{}} &
\textcolor{lightgray}{\xmark{}} &
opt. &
\cmark{} &
\href{https://github.com/Jiaoyang-Li/CBSH2-RTC}{\cmark{}} &
\cmark{} \\
\midrule
EECBS \cite{li_eecbs_2021} & G &
\textcolor{lightgray}{\xmark{}} &
\textcolor{lightgray}{\xmark{}} &
subopt. &
\cmark{} &
\href{https://github.com/Jiaoyang-Li/EECBS}{\cmark{}} &
\cmark{} \\
\midrule
LaCAM \cite{okumura_engineering_2024} & G &
\textcolor{lightgray}{\xmark{}} &
\textcolor{lightgray}{\xmark{}} &
subopt. &
\cmark{} &
\href{https://github.com/kei18/lacam0}{\cmark{}} &
\cmark{} \\
\midrule
MAPF-LNS2 \cite{li_mapf-lns2_2022} & G &
\textcolor{lightgray}{\xmark{}} &
\textcolor{lightgray}{\xmark{}} &
subopt. &
\cmark{} &
\href{https://github.com/Jiaoyang-Li/MAPF-LNS2}{\cmark{}} &
\cmark{} \\
\midrule
SCRIMP \cite{wang_scrimp_2023} & G &
\textcolor{lightgray}{\xmark{}} &
\textcolor{lightgray}{\xmark{}} &
subopt. &
\cmark{} &
\href{https://github.com/marmotlab/SCRIMP}{\cmark{}} &
\cmark{} \\
\midrule
MAPF-GPT-DDG \cite{andreychuk_advancing_2025} & G &
\textcolor{lightgray}{\xmark{}} &
\textcolor{lightgray}{\xmark{}} &
subopt. &
\cmark{} &
\href{https://github.com/Cognitive-AI-Systems/MAPF-GPT-DDG}{\cmark{}} &
\cmark{} \\
\midrule

CBS-MP \cite{solis_vidana_representation-optimal_2021} & R &
\textcolor{lightgray}{\xmark{}} & 
\cmark{} &
repr. & 
prob. &
\textcolor{lightgray}{\xmark{}} &
\textcolor{lightgray}{\xmark{}} \\
\midrule
CCBS \cite{andreychuk_multi-agent_2022} & R &
\textcolor{lightgray}{\xmark{}} & 
\textcolor{lightgray}{\xmark{}} &
* & 
* &
\href{https://github.com/PathPlanning/Continuous-CBS}{\cmark{}} &
\cmark{} \\
\midrule
PSIPP/CTC \cite{kasaura_prioritized_2022} & R &
\textcolor{lightgray}{\xmark{}} & 
\textcolor{lightgray}{\xmark{}} &
subopt. & 
\textcolor{lightgray}{\xmark{}} &
\href{https://github.com/omron-sinicx/PSIPP-CTC}{\cmark{}} &
\cmark{} \\
\midrule
CCBS+DK \cite{walker_clique_2024} & R &
\textcolor{lightgray}{\xmark{}} & 
\textcolor{lightgray}{\xmark{}} &
* & 
* &
\textcolor{lightgray}{\xmark{}} &
\textcolor{lightgray}{\xmark{}} \\
\midrule

SST* \cite{li_asymptotically_2016} & C &
\cmark{} & 
\cmark{} &
asym. &
prob. &
\href{https://ompl.kavrakilab.org/classompl_1_1control_1_1SST.html}{\cmark{}} &
\cmark{} \\
\midrule
MRRP \cite{binder_multi_2019} & C &
\textcolor{lightgray}{\xmark{}} & 
\cmark{} & 
subopt. & 
\textcolor{lightgray}{\xmark{}} &
\href{https://github.com/tuw-robotics/tuw_multi_robot}{\cmark{}} &
\textcolor{lightgray}{\xmark{}} \\
\midrule
K-CBS \cite{kottinger_conflict-based_2022} & C &
\cmark{} & \cmark{} &
- & prob. &
\href{https://github.com/aria-systems-group/Multi-Robot-OMPL}{\cmark{}} &
\textcolor{lightgray}{\xmark{}} \\
\midrule
SSSP \cite{okumura_quick_2023} & C &
\textcolor{lightgray}{\xmark{}} & 
\cmark{} &
- & prob. &
\href{https://github.com/Kei18/sssp}{\cmark{}} &
\textcolor{lightgray}{\xmark{}} \\
\midrule
db-CBS \cite{moldagalieva_db-cbs_2024} & C &
\cmark{} & 
\cmark{} &
asym. 
& prob. &
\href{https://github.com/IMRCLab/db-CBS}{\cmark{}} &
\cmark{} \\
\midrule
(E)CHPBS \cite{lin_multi-agent_2025} & C &
\cmark{} & 
\cmark{} &
subopt. & 
\textcolor{lightgray}{\xmark{}} &
\textcolor{lightgray}{\xmark{}} &
\textcolor{lightgray}{\xmark{}} \\
\bottomrule
\end{tabularx}

\par\smallskip
{\textsuperscript{\dag} All listed MAPF planners optimize the sum-of-costs (\textbf{SoC}). \\ 
\textbf{Dyn.}=agent dynamics; \textbf{Het.}=heterogeneous agents; \textbf{Opt.}=optimality; \textbf{Comp.}=completeness; 
\textbf{OSS}=open source; \textbf{Eval.}=evaluated in GRACE;
\textbf{repr.}=representation(optimal); \textbf{prob.}=probabilistically(complete); \textbf{asym.}=asymptotically(optimal);
\textbf{*}: ongoing discussions \cite{li_cbs_2025}; \textbf{-}: no proof.}
\endgroup
\end{table}

Most MAPF benchmarks traditionally evaluate search-based \cite{shen_tracking_2023, tan_benchmarking_2024, moghadam_guards_2024}, learning-based \cite{alkazzi_comprehensive_2024}, and reduction-based algorithms \cite{kaduri_experimental_2021, svancara_which_2024} in 2D grid environments with discrete time and uniform agents. Such benchmarks popularized by the MovingAI community \cite{sturtevant_benchmarks_2012} similarly emphasize grid-based MAPF, enabling standardized evaluation with simplified assumptions.

Recent work, including asprilo \cite{gebser_experimenting_2018}, SkyRover \cite{ma_skyrover_2025},  MRP-Bench \cite{schaefer_benchmark_2023}, REMROC \cite{heuer_benchmarking_2024}, has moved towards more expressive settings. These include intralogistic domains, real-life robotics use cases, human-shared environments, cross-domain coordination and platforms with agent size or kinematic constraints \cite{honig_path_2017}. As a realistic testbed in a continuous 2D environment, SMART \cite{yan_advancing_2025} leverages the robust discretized planning of the action dependency graph \cite{honig_persistent_2019}.

However, many benchmarks still assume grid-based path planners and environments. To address this, \cite{honig_trajectory_2018} introduces MAPF with generalized Conflicts (MAPF/C), which overcomes the grid limit by defining conflicts in continuous space. \cite{andreychuk_multi-agent_2022} proposes a continuous-time MAPF benchmark that works on both grids and roadmaps. These capture complexities such as continuous travel times, heterogeneous agents, and arbitrarily placed objects. In kinodynamic MRMP, more realistic comparisons are offered, although often under different assumptions \cite{moldagalieva_db-cbs_2024}. In other related domains, such as multi-goal multi-robot path planning in continuous space, mrmg-planning \cite{hartmann_benchmark_2025} formulates a single robot path planning approach to solve multi-robot problem, facing scalability issue. Our work builds on these efforts to support the rigorous evaluation of diverse agents across three environmental representations: grid-based environments, roadmap graphs, and shared continuous spaces.

Beyond static benchmarks, community challenges and competitions have further driven advancements in the research field. The League of Robot Runners (LoRR) provides standardized MAPF competitions with new domains (e.g., fulfillment, sortation) \cite{chan_league_2024}. Flatland targets large-scale railway traffic on grid graphs \cite{mohanty_flatland-rl_2020}. At the system level, RoboCup Logistics League (RCLL) tackles smart-factory intralogistics with real robots, evaluating far beyond 2D planning alone \cite{barros_2025_robocup}. GRACE complements these efforts by isolating and unifying the multi-robot planning layer across grid/roadmap/continuous representations, while remaining compatible with grid-style challenge maps (e.g., LoRR/Flatland) and clearly scoping out hardware-centric aspects of RCLL.

We summarize the most widely used MAPF and MRMP simulation benchmarks in Tab.~\ref{tab:benchmark}.

\subsection{Planner Families Across Representations}
\label{existing_algorithms}
A detailed comparison of 2D MAPF and MRMP solvers is shown in Tab. \ref{tab:algorithm}. A concise summary of the main planning families is reported below.
\subsubsection{Grid-centric MAPF (Rep-G)} On occupancy grids with discrete time, MAPF planners extensively exploit graph structure and uniform moves. Conflict-Based optimal and suboptimal Search (CBS) minimize SoC under vertex / edge conflict constraints \cite{sharon_conflict-based_2015, li_pairwise_2021, li_eecbs_2021, chan_flex_2022}. For scalability to larger problem instances, methods like Large-Neighborhood Search (LNS) and related metaheuristics trade optimality for computational efficiency \cite{li_mapf-lns2_2022, chan_anytime_2024}. Priority-based schemes offer strong speed with weaker optimality and completeness guarantees \cite{chan_greedy_2023, okumura_lightweight_2025}. Recently, learning-driven approaches, including decentralized MARL policies \cite{sartoretti_primal_2019, veerapaneni_improving_2024}, and search guided by learned heuristics \cite{tang_himap_2024}, aim to improve responsiveness in dense regimes. Furthermore, hybrid models combine learned proposals with classical repair \cite{wang_lns2rl_2025}.

\subsubsection{Roadmap-based MAPF (Rep-R)} When continuous environments are abstracted to directed graphs, timing becomes non-uniform and continuous. Safe Interval Path Planning (SIPP) \cite{phillips_sipp_2011} underpins Continuous Conflict-Based Search (CCBS) variants \cite{andreychuk_multi-agent_2022} and prioritized formulations with precomputed conflict intervals, scaling to large teams \cite{kasaura_prioritized_2022}. Most roadmap methods assume disc robots with single-integrator kinematics, enabling richer timing than grids without full kinodynamics.

\subsubsection{Continuous-space MAMP/MRMP (Rep-C)} Multi-Agent Motion Planning (MAMP) generalizes MAPF to continuous spaces and agent dynamics. Hybrid frameworks bridge MAPF and MAMP: MAPF-POST \cite{honig_multi-agent_2016} refines discrete plans into temporally coordinated trajectories. Other approaches combine a discrete search with continuous sampling \cite{okumura_quick_2023}. Fully continuous MRMP operates directly in continuous state–time with footprints and dynamics. Sampling-based planners address feasibility under differential constraints \cite{shome_2020_drrt, dayan_2023_near}. Recent kinodynamic MRMP \cite{kottinger_conflict-based_2022-1, moldagalieva_db-cbs_2024} combines discrete conflict reasoning with dynamic feasibility checks, offering a logical middle ground. 

\section{Problem Formulation} \label{sec:problem}

This section formalizes the multi-agent planning problem within GRACE, outlining its foundational environment, agent, and planner models (Sec.\ref{sec:env_model_ps}–\ref{sec:planner_ps}). The goal is to find feasible (Sec.\ref{sec:feasibility_ps}) per-agent plans from specified start-goal pairs, adhering to these models. A summarizing table (Table \ref{tab:featuremap}) at the end of this section provides an overview of key relationships, with all symbols and specialized terminology formally defined in Sec.\ref{sec:env_model_ps}–\ref{sec:assumption_ps}.

\subsection{Environment Model and Agent's Goals} \label{sec:env_model_ps}

Let the workspace be a closed, bounded 2D region $\Omega \subset \mathbb{R}^2$ with polygonal obstacles $\mathcal{O} \subset \Omega$. The free space is defined as $X_{\mathrm{free}} = \Omega \setminus \mathcal{O}$. GRACE employs abstraction operators $\Phi_{\text{env}}^{\rho}(\Omega,\mathcal{O})$, \(\rho\in\{\text{cont},\text{road},\text{grid}\}\), to transform the detailed workspace into simplified representations. These operators produce three levels of abstraction:

\begin{itemize}
    \item \textbf{Continuous}: The environment remains \(X_{\mathrm{free}}\) (“as is”); goals ($g_i$ herein) are locations in continuous space. Thus, $\Phi_{\text{env}}^{\text{cont}}(\Omega,\mathcal{O})=X_{\mathrm{free}}$.
    \item \textbf{Roadmap}: A directed graph \(G_r=(V_r,E_r)\) is extracted from \(X_{\mathrm{free}}\); goals are mapped to vertices in \(V_r\). Thus, $\Phi_{\text{env}}^{\text{road}}(\Omega,\mathcal{O})=G_r$.
    \item \textbf{Grid}: An \(M\)-connected occupancy grid \(G_g=(V_g,E_g)\) discretizes \(\Omega\); goals snap to containing cells. Thus, $\Phi_{\text{env}}^{\text{grid}}(\Omega,\mathcal{O})=G_g$.
\end{itemize}

\subsection{Agent Model} \label{sec:agent_ps}
Let $N$ be the total number of agents. Each agent \(\{i\}_1^N\) has a configuration space \(Q_i \in \{\mathbb{R}^2, \mathrm{SE}(2)\}\), corresponding to translation-only or full planar rigid-body pose (with \(\mathrm{SE}(2) = \mathbb{R}^2 \times \mathrm{SO}(2)\)). Its properties include a start location \(s_i\in Q_i\), a convex footprint \(S_i\subset\mathbb{R}^2\), and an action model specifying its discrete, kinematic or kinodynamic motion profile \(f_i\).

Let \(B(c,r)=\{z\in\mathbb{R}^2:\|z-c\|\le r\}\) be the closed Euclidean ball with center $c$ and radius $r$.
For a compact robot footprint \(S_i\) (in its body frame), we define the minimum enclosing circle
\[
c_{\mathrm{eff}}(S_i), r_{\mathrm{eff}}(S_i)\;:=\;\operatorname*{arg min}_{c\in\mathbb{R}^2}\ \max_{z\in S_i}\|z-c\|
\]
with center $c_{\mathrm{eff}}$ and radius $r_{\mathrm{eff}}$. To compare heterogeneous footprints with a single-disc model, we define the minimal common effective radius $r_{\mathrm{com}} \;:=\; \max_{i=1,\dots,N} r_{\mathrm{eff}}(S_i)$, that is, the smallest \(r\) such that \(S_i \subseteq B(0,r)\) for all agents.

We take \(c_{\mathrm{eff}}(S_i)\) as the body-fixed reference point and denote its world position by \(p_i(t)\in\mathbb{R}^2\) (so if \(Q_i=\mathrm{SE}(2)\), the agent's state is $x_i(t) = (p_i(t),\theta_i(t))$).

\begin{itemize}
    \item \textbf{Continuous}: Agents retain their arbitrary convex footprints \(S_i\) and heterogeneous motion profile \(f_i\). Thus, we define \(\Phi_{\text{ag}}^{\text{cont}}\) as the identity (no abstraction). 

    \item \textbf{Roadmap}: For comparability, agents are abstracted to \(r_{\mathrm{com}}\) with stop–go single-integrator:
    \(\Phi_{\text{ag}}^{\text{road}}(S_i,f_i)=(r_{\mathrm{com}},\,\dot x_i=u_i,\, \|u_i^\text{linear}\|\in\{0,\nu\})\),
    where \(\nu > 0\) is the common linear speed.

    \item \textbf{Grid}: Agents are abstracted to homogeneous, tile-sized discs of radius \(r_{\mathrm{com}}\) with motion defined by \(M\)-connected moves with discrete unit time steps: $\Phi_{\text{ag}}^{\text{grid}}(S_i,f_i)=(\mathrm{cell},\,\mathcal{A}_{M\text{-conn}}\cup\{(0,0)\},\,\Delta t)$. 
    Here, \(\mathcal{A}_{M\text{-conn}}\subset\mathbb{Z}^2\) denotes the allowed neighbor offsets (e.g., for 4-connectivity), \((0,0)\) encodes the “wait” action, and \(\Delta t > 0\) is the discrete time step (often \(1\)).
    The grid cell side length \(a(M)\) is determined by the maximum displacement allowed by \(M\)-connectivity $L_{\max}(M):=\max_{d\in\mathcal{A}_{M\text{-conn}}}\|d\|_2$ such that \(a(M)=2\,L_{\max}(M)\,r_{\mathrm{com}}\). For example, \(a(4)=2r_{\mathrm{com}}\) and \(a(8)=2\sqrt{2}\,r_{\mathrm{com}}\).
\end{itemize}

\subsection{Planner Model}\label{sec:planner_ps}

Let \(\mathcal{P}_\rho\) be a registry of multi-agent planners compatible with representation \(\rho\in\{\text{cont},\text{road},\text{grid}\}\). Given an environment representation, agent abstractions, and start–goal pairs \(\{(s_i,g_i)\}_{i=1}^{N}\), if successful, a planner \(P\in\mathcal{P}_\rho\) returns plans \(\Pi=\{\pi_i\}_{i=1}^{N}\) for each agent:

\begin{itemize}
    \item \textbf{Continuous}: \(\pi_i\) is a (kino)dynamically feasible trajectory (into \(X_{\mathrm{free}}\) in its position component).
    \item \textbf{Roadmap}: \(\pi_i\) is a time-parameterized path along nodes/edges of \(G_r\) (continuous time).
    \item \textbf{Grid}: \(\pi_i\) is a sequence of time-stamped cell moves on \(G_g\) (discrete time).
\end{itemize}


\subsection{Feasibility}\label{sec:feasibility_ps}

A plan $\pi_i:[0,T_i]\!\to\! Q_i$ with a finite horizon \(T_i\), is considered feasible if it satisfies all constraints related to obstacle avoidance and inter-agent collision prevention, as defined by the chosen representation.

\begin{itemize}
    \item \textbf{Continuous}: Let \(R(\theta)\) be the planar rotation by angle \(\theta\), and let $\oplus$ be the Minkowski sum operator. The occupied region of agent \(i\) at time \(t\) is
    \[
    X_i(t)=
    \begin{cases}
      p_i(t)\oplus S_i, & \text{if } Q_i=\mathbb{R}^2,\\
      p_i(t)\oplus \big(R(\theta_i(t))\,S_i\big), & \text{if } Q_i=\mathrm{SE}(2).
    \end{cases}
    \]
    A plan is feasible iff \(X_i(t)\subseteq X_{\mathrm{free}}\) for all $i$ and \(\forall t\in[0,T_i]\) and \(X_i(t)\cap X_j(t)=\varnothing\) for all \(i\neq j\) and \(\forall t\); additionally, \(\pi_i\) satisfies \(f_i\).
    \item \textbf{Roadmap}: We enforce space–time feasibility for discs of radius \(r_{\mathrm{com}}\). Feasibility requires \(\|p_i(t)-p_j(t)\|\ge 2\,r_{\mathrm{com}}\) for all \(t\), enforced via (i) vertex capacity~1, (ii) no head-on edge swaps, (iii) same-edge headway \(\ge 2\,r_{\mathrm{com}}/\nu\), and (iv) treating geometric edge intersections as virtual vertices with capacity~1.
    \item \textbf{Grid}: With cell side \(a(M)=2\,L_{\max}(M)\,r_{\mathrm{com}}\) and linear interpolation over \(\Delta t\), we forbid co-occupancy, opposing edge swaps \cite{stern_multi-agent_2019}, diagonal crossings, and corner-meets; these imply \(\|p_i(t)-p_j(t)\|\ge 2\,r_{\mathrm{com}}\) throughout each unit move.

\end{itemize}
Let \(\Pi_{\text{feasible}}\) denote the subset of feasible plans.

\subsection{Objectives} \label{sec:objectives}

While feasibility is paramount, GRACE also supports optimization for common objectives. These include, but are not limited to, makespan and SoC:
\[
\min_{\Pi_{\text{feasible}}} \ J(\Pi), \text{with } J\in\{\text{makespan},\ \text{SoC},\ \dots\}.
\]
The framework does not fix \(J\); metrics are computed consistently across representations.

\begin{table}[t]
\caption{Feature mapping.}
\label{tab:featuremap}
\centering
\begin{tabular}{lccc}
\toprule
\textbf{Aspect} & \textbf{Continuous} & \textbf{Roadmap} & \textbf{Grid} \\
\midrule
Time        & continuous          & continuous           & discretized \\
Footprint   & \(S_i\) (convex)    & disc \(r_{\mathrm{com}}\) & disc \(r_{\mathrm{com}}\) \\
Orient.     & if \(Q_i{=}\mathrm{SE}(2)\)   & --                  & -- \\
Kinematics  & \(f_i\)                       & SI(\(\nu\))           & unit moves \\
Collision   & swept \(S_i\)                 & cap/edges/ints      & MAPF rules \\
Goals       & invariant set                 & remain at node      & remain at cell \\
\bottomrule
\end{tabular}
\par\smallskip
SI=single-integrator; cap=capacity; ints=geometric intersections. Refer to Sec. \ref{sec:problem} for definitions of concepts and symbols.

\end{table}

\subsection{Assumptions and Scope}\label{sec:assumption_ps}
The problem formulation operates under the following assumptions: environments are static and fully known; tasks are single-shot with fixed start-goal pairs. All agent shapes are fixed (no articulated robots) and convex. Motion is constrained to a planar domain; orientation is included where relevant. A parked-at-goal model is used: upon arrival at time \(T_i\), agents remain at their goal for all \(t\ge T_i\). The choice of representation is user-controlled; GRACE provides reproducible mappings and consistent instrumentation for all abstraction levels.

\subsection{Problem statement} Given a base world \((\Omega,\mathcal{O},\{S_i,f_i,(s_i,g_i)\}_{i=1}^N)\) and a representation \(\rho\in\{\text{cont},\text{road},\text{grid}\}\), GRACE instantiates a planner-ready problem via \(\Phi_{\text{env}}^{\rho}\) and \(\Phi_{\text{ag}}^{\rho}\), then re-embeds returned plans to compute common metrics. Cross-representation comparisons are restricted to tasks that admit multiple \(\rho\) (the overlapping regime); outside this regime we report representation-specific trade-offs rather than excluding such cases.

\section{GRACE Architecture}

\begin{figure}[t]
\centering
\includegraphics[width=\columnwidth]{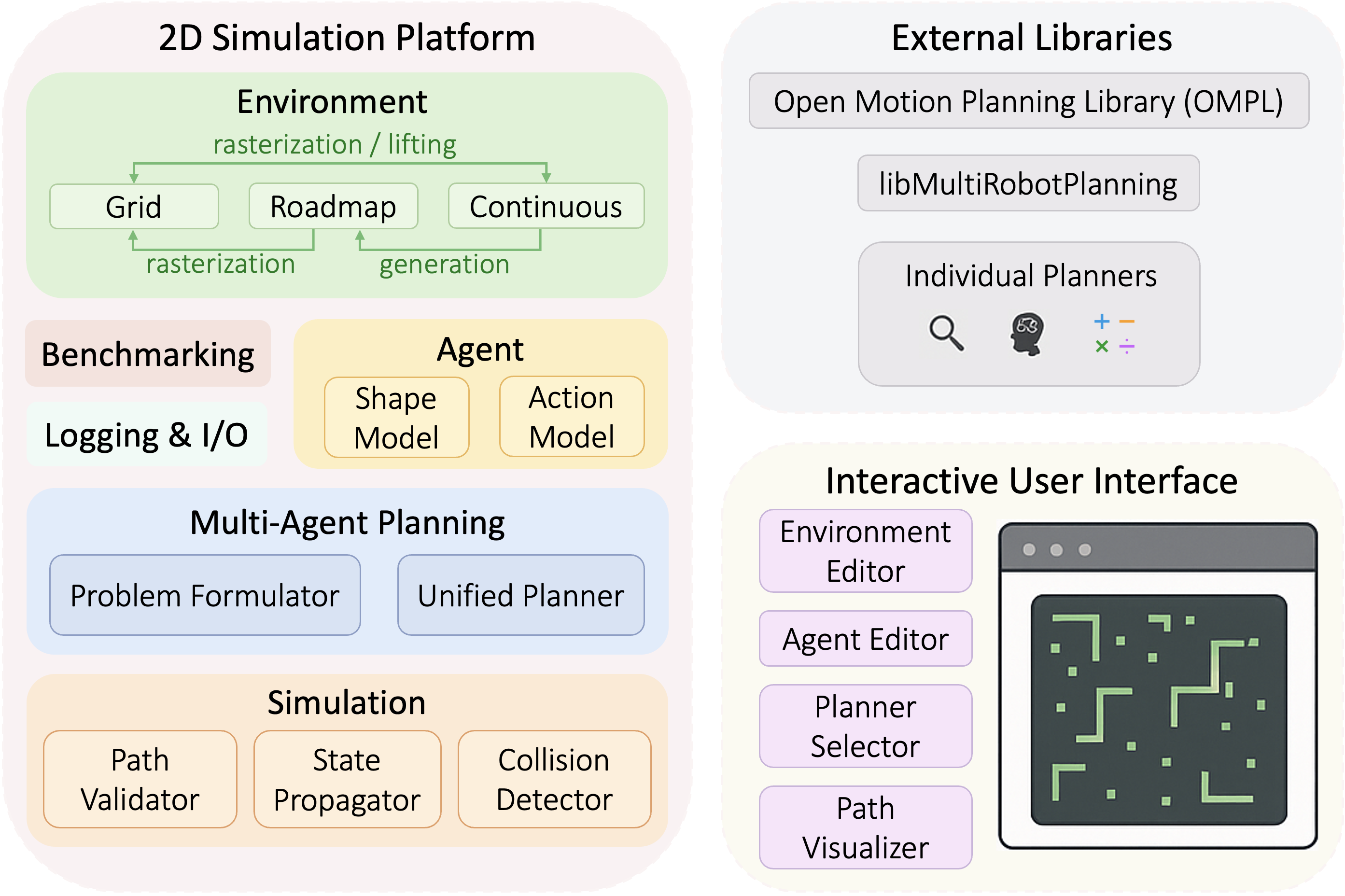}
\caption{Architectural Diagram of GRACE.}
\label{fig2}
\end{figure}

Fig. ~\ref{fig2} provides an overview of GRACE's architecture: a C++20 core based on Box2D \cite{box2dBox2D} for deterministic, continuous-time simulation, an interactive UI (SFML~+~ImGui), and external planners connected via a unified interface. A single physical scenario can be realized at multiple abstraction levels, planned by different algorithms, and executed under one simulator.

\subsection{Representation Operators and Consistency} \label{sec:env_arc}

We implement the environment and agent abstractions from Sec.~\ref{sec:env_model_ps} and \ref{sec:agent_ps} with explicit, reproducible parameters, so cross-representation studies do not require re-authoring.

\textit{Continuous \(\rightarrow\) Roadmap.} Given \(X_{\mathrm{free}}=\Phi_{\text{env}}^{\text{cont}}(\Omega,\mathcal{O})\), we construct a directed roadmap \(G_r=(V_r,E_r)\) using OMPL \cite{sucan2012open} and pin starts/goals \(\{(s_i,g_i)\}\) to \(V_r\). Edge weights store Euclidean length \(\ell_e\); with the common speed \(v>0\) from \(\Phi_{\text{ag}}^{\text{road}}\), traversal times are \(\tau_e=\ell_e/v\).

\textit{Continuous \(\rightarrow\) Grid.} We rasterize \(\Omega\) with cell side \(a(M)=2\,L_{\max}(M)\,r_{\mathrm{com}}\). Occupancy is computed by polygon coverage of obstacles. We use \(\Delta t=1\).

\textit{Roadmap \(\rightarrow\) Grid.} We rasterize edges of \(G_r\) onto the grid overlay to obtain a graph-constrained grid that preserves roadmap topology and  enables discrete-time scheduling.

\textit{Grid \(\rightarrow\) Continuous.} We embed polylines through cell centers in world coordinates and reconstruct obstacles directly from the occupancy mask. 

\textit{Roadmap \(\rightarrow\) Continuous.} 
We inflate obstacles per robot by the disk surrogate \(B(0,r_{\mathrm{eff}}(S_i))\):
\[
\mathcal{O}^{(i)}=\mathcal{O}\oplus B(0,r_{\mathrm{eff}}(S_i)),\qquad
X_{\mathrm{free}}^{(i)}=\Omega\setminus \mathcal{O}^{(i)}.
\]
Given polyline vertices \((p_k)_{k=0}^{K}\) defining robot $i$'s path on the roadmap, each path segment is validated obstacle-free passage via the swept set:
\[
[p_k,p_{k+1}]\oplus B(0,r_{\mathrm{eff}}(S_i))\ \subseteq\ X_{\mathrm{free}}^{(i)}.
\]

\subsection{Multi-agent Planning Pipeline}

\textit{Problem Formulator.} Given \(\rho\in\{\text{cont},\text{road},\text{grid}\}\), the formulator compiles planner-ready instances that respect the time model, capacity/avoidance rules (Sec.~\ref{sec:feasibility_ps}), and objective hooks (Sec.~\ref{sec:objectives}). It supplies graphs/costs for roadmap/grid and collision predicates/state propagators for continuous space.

\textit{Unified Planner} Our unified planner normalizes inputs (instance, budgets) and outputs (paths/trajectories with timestamps, status, and costs) across grid MAPF, roadmap MAPF, and continuous MRMP planners. Planners run as shared libraries or subprocesses with wall-clock/node budgets. 

\subsection{Simulation Core} \label{simulation_core}

Box2D serves as the cross-platform deterministic, continuous-time substrate. We advance at \(60\)~Hz (\(\Delta t_{\mathrm{sim}}=1/60\)~s) with substepping (default \(n=4\)) for contact stability. Each agent tracks its reference trajectory kinematically; collision checking uses broad-phase dynamic AABB trees and narrow-phase contact manifolds. We log contact events, nearest-approach distances, and tracking error. Given the same instance, seeds, and step parameters, the state trace is bitwise identical.

\subsection{Interactive User Interface}

The UI supports (i) environment edits (obstacles, resolution/roadmap parameters), (ii) agent edits (footprints, dynamics, starts/goals), (iii) planner/budget selection, and (iv) plan/execution visualization. Overlays include occupancy/roadmap layers, time-expanded reservations, clearance heatmaps, and per-agent timelines. Edits are transactional: dependent artifacts are invalidated and recomputed so UI and headless runs remain consistent.

\section{Experiments and Results}

We evaluate GRACE against three claims: generality, its capacity to yield new research insights, and practicality. 

\textit{Experimental Conditions.} All studies are conducted headless with fixed seeds and recorded manifests to ensure reproducibility. Grid experiments always use 4-connected moves; roadmap experiments use disk agents with single-integrator dynamic; continuous experiments execute time-parameterized trajectories with collision checks in Box2D. All experiments are performed on a laptop with an Intel(R) Core(TM) i7-13850HX CPU and 32GB of memory. For learning-based grid planners, we download trained models from official repositories and evaluate inference-time only. We leverage GPU on the same laptop with NVIDIA RTX 3500 Ada (12 GB).

\textit{Planners.} Tab.~\ref{tab:algorithm} lists the planners and representations they operate on and whether we evaluate them in GRACE (\textbf{Eval.}). We integrate official implementations via shared libraries or subprocesses (links in \textbf{OSS}). All run under identical instance definitions and budgets: 1\,min (grid/roadmap) and 5\,min (continuous) with default hyperparameters.

\textit{Metrics.} We implement various metrics to support the benchmark, including success rate (\%), SoC (path length in meter), makespan (s), and planning time (s). In addition, we also record the simulation speed, quantified as the Real-Time Factor (RTF), which is the ratio of simulated time to wall time (RTF = simulated time / wall time).

\subsection{Cross-Environment-Model Benchmarking}\label{sec:crossenv_sim}

\begin{figure*}[h]
\centering
\includegraphics[width=2\columnwidth]{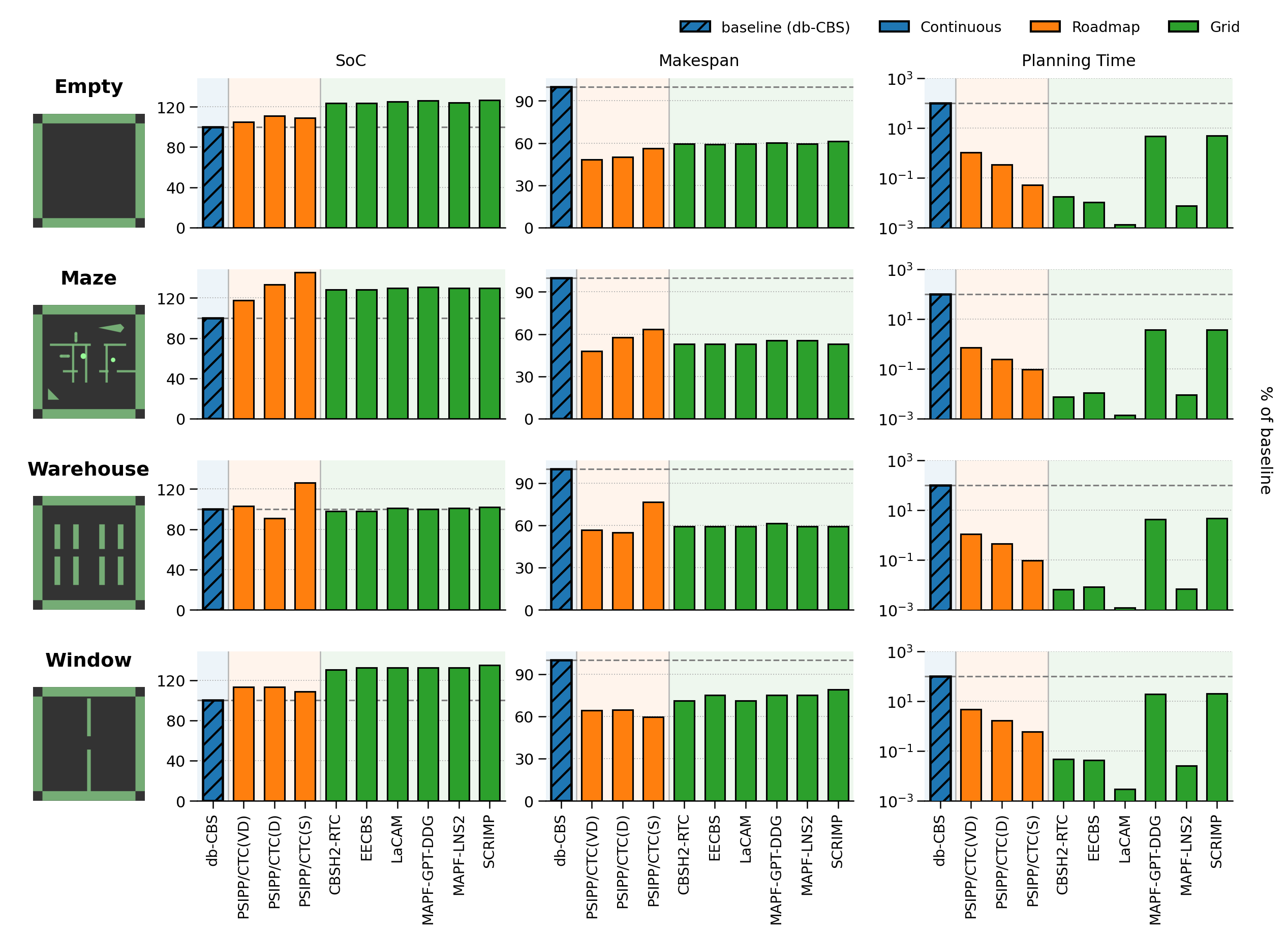}
\caption{Cross-Environment Comparison of MRMP and MAPF Solvers: SoC, Makespan, Planning Time (db-CBS=100\%)}
\label{fig3}
\end{figure*}

\textit{Purpose.} We assess \textit{generality} by expressing a fixed task set at multiple abstraction levels, including continuous (MRMP), roadmap (MAPF), and grid (MAPF). We measure how each representation trades SoC, makespan, and planning time.

\textit{Setup.}
We use a \(5{\times}5\,\mathrm{m}^2\) workspace with four map families: \texttt{empty}, \texttt{maze}, \texttt{warehouse}, and \texttt{window}. Teams contain up to \(N{=}8\) robots with heterogeneous convex footprints (maximum effective radius \(r_{\mathrm{eff}}^{\max}{=}0.08\,\mathrm{m}\)). To isolate representation effects, we use identical double-integrator bounds for continuous representation: \(v\in[-0.5,0.5]\,\mathrm{m/s}\), \(a\in[-2,2]\,\mathrm{m/s^2}\), consistent with \cite{moldagalieva_db-cbs_2024}. Roadmaps are instantiated at three densities (S=sparse, D=dense, SD=superdense); grids use cell size derived from \(r_{\mathrm{com}}\) (\(32{\times}32\)).

\textit{Results.}
As shown in Fig. \ref{fig3}, we observe clear representation-induced trade-offs. Roadmap consistently achieves near-continuous SoC with shorter makespans and much faster planning; higher roadmap density tightens SoC and makespan at a small runtime cost. Grid further accelerates planning and often matches roadmap-level makespans, but incurs a larger SoC penalty. Makespan across MRMP and MAPF is \emph{not} directly comparable because the time/kinematic models differ (double-integrator vs. single-integrator/unit-step); thus, makespan values should be read qualitatively and relative to the per-map continuous baseline.

\textit{Observations.}
(1) Roadmap is “good enough” in these tasks. Across maps, roadmap SoC stays within \(\approx\)108–118\% of continuous MRMP, while makespan drops to \(\approx\)52–61\% and planning time to \(\approx\)0.16–1.6\% of db-CBS.
(2) Grid trades more SoC for similar makespan and extreme speed. Grid averages \(\approx\)123\% SoC, \(\approx\)61\% makespan, and \(\approx\)2.4\% planning time.
(3) Density matters on roadmaps. Superdense roadmaps tighten SoC (108\%) and makespan (52\%) at modest runtime (1.6\%), indicating a controllable speed–quality knob.

\textit{Takeaway.} Moving from continuous to roadmap trades kinodynamic fidelity for speed, yielding faster planning and shorter makespans with a small SoC penalty. Further abstraction to grids amplifies planning speed but often increases makespan and SoC. However, environmental structure (e.g., aisles) can mitigate these penalties, sometimes favoring discrete abstractions. These findings underscore GRACE's ability to expose representation-induced trade-offs, revealing when discrete abstractions are adequate and when kinodynamic fidelity becomes critical. 

\subsection{Cross-MAPF-Planner Benchmarking on Grids}\label{sec:crossplanner_sim}

\textit{Purpose.} We evaluate GRACE's value in \textit{facilitating rigorous, comparative analysis} by demonstrating its capability as a reproducible, drop-in grid benchmark also for {single-representation} (e.g., grid-only) studies. This test shows direct compatibility with established MAPF community evaluations, while preserving our unified protocol for instance specification, budgets, logging, and evaluation.

\textit{Setup.} We evaluate six planners on four canonical grid maps (\texttt{random-64-64-10}, \texttt{maze-32-32-4}, \texttt{warehouse-10-20-10-2-1}, \texttt{den312d}) with 25 scenarios each, all from \cite{sturtevant_benchmarks_2012}. The tile side length is 1 meter. Agent counts ($N$) are swept, applying fixed wall-clock budgets and the standard success criterion (Sec.~\ref{sec:feasibility_ps}).

\begin{figure}[]
\centering
\includegraphics[width=\columnwidth]{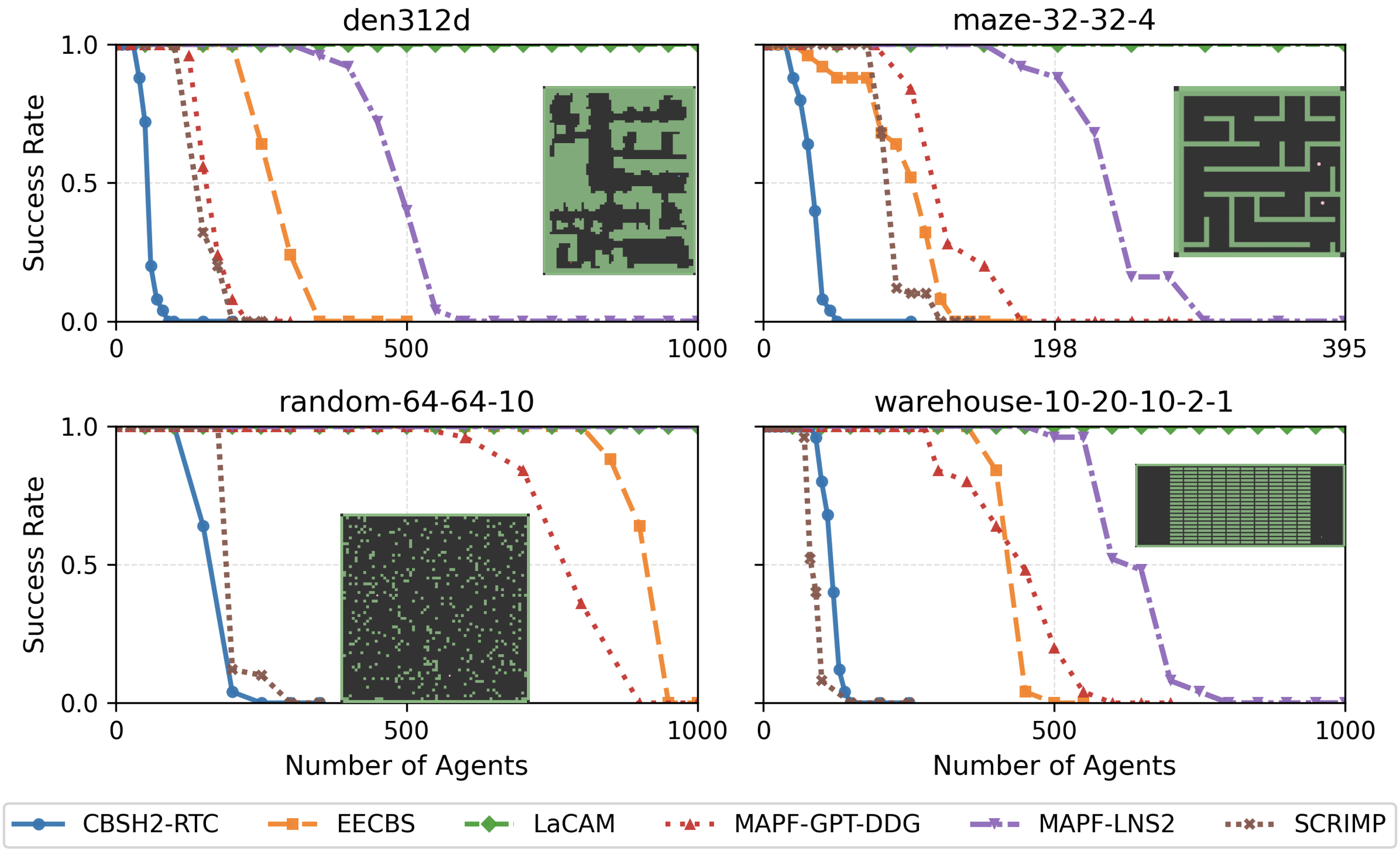}
\caption{Grid MAPF Comparison: Success Rate with Number of Agents.}
\label{fig:grid_success_vs_n}
\end{figure}

\textit{Success Rate.} Fig.~\ref{fig:grid_success_vs_n} reports success rate vs.\ \(N\). Three consistent trends: (i) LaCAM achieves \(100\%\) success across the entire sweep on all four maps in our setting; (ii) EECBS is strongest at low–mid \(N\) but degrades earlier on \texttt{den312d/maze-32-32-4}, while MAPF-LNS2 degrades more gradually at larger \(N\); (iii) learned/hybrid methods (MAPF-GPT-DDG, SCRIMP) can match or exceed search on random/warehouse but show higher across-map variance.

\textit{SoC and Planning Time.} To compare path quality fairly, we compute average SoC and planning time only on the subset of instances where \emph{all} planners succeed. CBSH2-RTC achieves the lowest SoC, followed by EECBS; LNS and learned/hybrid trade slightly higher SoC for improved scale tolerance; LaCAM prioritizes robustness at larger \(N\).

\begin{table}[t]
\caption{SoC and planning time averaged over the common-success subset across all planners.}
\label{tab:grid_soc_common}
\centering\scriptsize
\begin{tabular}{lcc}
\toprule
Planner & Mean SoC (m) $\downarrow$ & Mean planning time (s) $\downarrow$ \\
\midrule
CBSH2\textendash RTC & \textbf{3827 $\pm$ \textcolor{gray}{286}} & 4.61 $\pm$ \textcolor{gray}{1.84} \\
EECBS              & 3940 $\pm$ \textcolor{gray}{288}   & 0.63 $\pm$ \textcolor{gray}{0.08} \\
MAPF\textendash LNS2 & 4206 $\pm$ \textcolor{gray}{321}  & 0.45 $\pm$ \textcolor{gray}{0.05} \\
MAPF\textendash GPT\textendash DDG & 4257 $\pm$ \textcolor{gray}{333}   & 5.66 $\pm$ \textcolor{gray}{0.26} \textsuperscript{\dag} \\
LaCAM         & 4511 $\pm$ \textcolor{gray}{358}  & \textbf{0.01 $\pm$ \textcolor{gray}{0.00}} \\
SCRIMP        & 5290 $\pm$ \textcolor{gray}{422}    & 20.86 $\pm$ \textcolor{gray}{1.56} \textsuperscript{\dag} \\
\bottomrule
\end{tabular}
\par\smallskip
\textsuperscript{\dag}\, Includes Python driver and model loading overhead (Docker). \\
\textcolor{gray}{gray}: 95 \% Confidence Interval (CI).

\end{table}

\textit{Observation and follow-up.} Because LaCAM already achieves \(100\%\) success on all four maps in the nominal sweep, we run an additional large-scale stress test, as shown in Sec. \ref{subsec:scalability}; LaCAM maintained \(100\%\) success until \(N=2500\) in our 1 minute budget in that extended regime.

\textit{Takeaway.} This grid-only benchmark confirms GRACE's utility as a reproducible platform for MAPF planner evaluation, fully compatible with existing community standards. The results further illuminate the trade-offs between planner robustness, optimality, and efficiency across map types and agent densities within GRACE's unified framework.

\subsection{Scalability and Software Resources} \label{subsec:scalability}

\textit{Purpose.} We evaluate \textit{practicality} through the runtime scalability, resource usage, and execution determinism of GRACE’s simulator and benchmarking stack, and we report capacity-style metrics that make comparisons across hardware and existing simulators straightforward.

\textit{Setup.} We use the same integration settings as Sec.~\ref{simulation_core}. For each run, we record OS-reported peak CPU utilization and process resident peak memory. To facilitate cross-paper comparison, we summarize (i) the largest team size with a real-time factor $\mathrm{RTF}\!\geq\!1$, (ii) the team size at which the median curve crosses a reference level $\mathrm{RTF}\!=\!10$ (as used in \cite{yan_advancing_2025}), and (iii) peak-memory scaling via a linear fit; we report the slope in MB/agent and coefficient of determination $R^2$. Collision checking remains enabled.

\textit{Results.} We stress-test the grid representation by replaying LaCAM plans on \texttt{warehouse\_large} (size: 540m $\times$ 140m) with team sizes up to $N\!=\!2500$. GRACE sustains real time up to the largest tested team size at $N\!=\!2500$ and remained above the reference level at approximately $N\!=\!550$. Peak memory grows linearly with team size; a least-squares fit yielded a slope of $\sim\!0.179$\,MB/agent with $R^2\!=\!1.00$, indicating a small per-agent footprint and modest constant overhead. Peak CPU stabilizes around $100$–$110\%$ in this configuration, consistent with a lightly threaded but memory-efficient workload. Table~\ref{tab:largescale_keypoints} reports representative points on the median curves.

\begin{table}[t]
\caption{Key points across $N$ during path-playback on grid representation \texttt{warehouse\_large} using LaCAM.}
\label{tab:largescale_keypoints}
\centering
\scriptsize
\begin{tabular}{rrrr}
\toprule
$N$ & RTF (median) & Peak CPU (\%) & Peak Mem (GB) \\
\midrule
2    & 1288.21 & ~9.9  & 0.039 \\
550  &   10.59 & 108.6 & 0.134 \\
1000 & 5.41 & 107.4 & 0.213 \\
2000 & 2.39 & 107.1 & 0.387 \\
2500 &    1.84 & 107.2 & 0.476 \\
\bottomrule
\end{tabular}
\par\smallskip
 We report median Real-Time Factor (RTF), peak CPU, and peak memory at selected team sizes. Peak CPU denotes the
OS-aggregated utilization during playback and indicates
system load rather than hardware-independent cost.
\end{table}

GRACE uses the same waypoint-tracking and contact model for all environment representations (Sec. \ref{simulation_core}), so differences observed come from the plans and not from execution.

\section{Conclusion}

We present GRACE, a unified simulator and benchmark that enables commensurate evaluation of discrete MAPF and kinodynamically faithful MAMP across grid, roadmap, and continuous representations under a single protocol. Experiments verify our claims of generality (convertible environments; heterogeneous agents and obstacles), usefulness (fair, reproducible comparisons that expose representation–algorithm trade-offs), and practicality (fast-speed playback and deterministic logs on modest hardware).

Limitations include a 2D focus, static maps, fixed convex geometries, and agent abstractions for roadmap/grid. Future work will explore uncertainty and dynamic changes, lifelong/multi-goal settings, richer motion dynamics and objectives, automated abstraction/roadmap selection, and connectors to operational stacks, including ROS 2 and Fleet Management Systems (FMS), moving toward deployment.







\bibliographystyle{IEEEtran}
\bibliography{IEEEabrv,reference_small_modified}

\end{document}